\documentclass[10pt,twocolumn,letterpaper]{article}
\pdfoutput=1
\usepackage{times}
\usepackage{epsfig}
\usepackage{graphicx}
\usepackage{amsmath}
\usepackage{amssymb}
\usepackage{amsfonts}
\usepackage{booktabs}
\usepackage{siunitx}

\usepackage[pagebackref=true,breaklinks=true,letterpaper=true,colorlinks,bookmarks=false]{hyperref}

\begin{document}

%%%%%%%%% TITLE
\title{SnakeVoxFormer: Transformer-based Single Image\\Voxel Reconstruction with Run Length Encoding}

\author{Jae Joong Lee\\
Purdue University\\

\and
Bedrich Benes\\
Purdue University\\

}

\maketitle

%%%%%%%%% ABSTRACT
\begin{abstract}
Deep learning-based 3D object reconstruction has achieved unprecedented results. Among those, the transformer deep neural model showed outstanding performance in many applications of computer vision. We introduce SnakeVoxFormer, a novel, 3D object reconstruction in voxel space from a single image using the transformer. The input to SnakeVoxFormer is a 2D image, and the result is a 3D voxel model. The key novelty of our approach is in using the run-length encoding that traverses (like a snake) the voxel space and encodes wide spatial differences into a 1D structure that is suitable for transformer encoding. We then use dictionary encoding to convert the discovered RLE blocks into tokens that are used for the transformer. The 1D representation is a lossless 3D shape data compression method that converts to 1D data that use only about 1\% of the original data size. We show how different voxel traversing strategies affect the effect of encoding and reconstruction. We compare our method with the state-of-the-art for 3D voxel reconstruction from images and our method improves the state-of-the-art methods by at least 2.8\% and up to 19.8\%.
\end{abstract}

%%%%%%%%% BODY TEXT
\section{Introduction}
\begin{figure*}[hbt]
\centering
\includegraphics[width=0.95\linewidth]{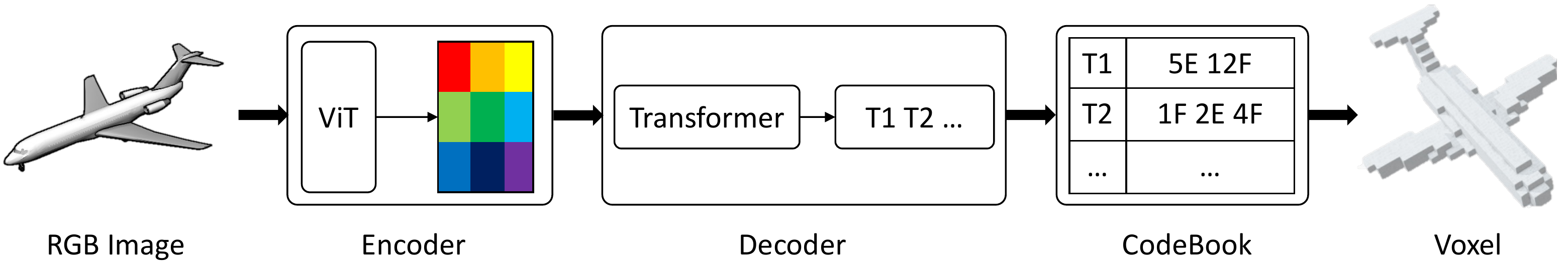}
  \caption{Overview of SnakeVoxFormer:
  The input is an RGB image that is encoded by using a ViT encoder to extract its features. The features are input to the transformer-based decoder that  produces tokens corresponding to image features. The tokens are stored in a codebook that allows the decompression of the Run Length Encoded data back to its original form in voxels.
  }\vspace{-3mm}
  \label{fig:overview}
\end{figure*}

3D object reconstruction from images is an important problem that has been addressed by many methods. In particular, captioning helps to explain images~\cite{Guo_2019_CVPR,vinyals2015show} or even videos~\cite{Hou_2019_ICCV,wu2017deep} in human language. One important area is voxel object reconstruction from single images that provides immediate space occupancy and data structure that is readily available for tasks that require collision detection. Transformer~\cite{attention} was introduced for Natural Language Process (NLP), and it has achieved supported an outstanding performance in many areas, such as image classification~\cite{vit, swin, Wang_2021_ICCV} and image captioning~\cite{ji2021improving,li2019entangled}. However, acquiring a 3D voxel model from a single image is an open problem that has been solved only partially.

We introduce SnakeVoxFormer, a novel algorithm for 3D voxel model reconstruction from a single image. The key idea of our approach is to encode the voxel space by traversing through it in a snake-like way. We encode all voxels by Run Length Encoding (RLE) compression~\cite{RLE} that stores the data as a sequence of couples of numbers~$[rep,val]$, where $rep$ indicates the number of repetitions and $val$ the voxel value, so e.g., the couple $[7,0]$ indicates seven consecutive empty voxels. We effectively convert the 3D voxel space into a 1D structure. In the next step, we create a dictionary of the detected structures (the codebase) that is then encoded as a set of tokens and used for learning with a transformer. For each input image, SnakeVoxFormer extracts image features by using Vision Transformer~\cite{vit} pre-trained on ImageNet~\cite{imagenet} and maps them to the corresponding tokens. 

SnakeVoxFormer outperforms the state-of-the-art methods for 3D voxel reconstruction
3D-R2N2~\cite{3dr2n2}, AttSets~\cite{yang2020robust}, Pix2Vox-A~\cite{pix2vox}, Pix2Vox++/A~\cite{pix2voxpp}, VoIT~\cite{wang2021multi}, VoIT+~\cite{wang2021multi}, EVoIT~\cite{wang2021multi}, and TMVNet~\cite{TMVNet} by at least 2.8\% and up to 19.8\%.

We claim the following contributions: 1)~Compressing voxels by Run Length Encoding and dictionary-based encoding into 1D data representation suitable for deep learning, 2)~transformer-based 3D object reconstruction network with multiple views using RLE. 

We will provide the source code of SnakeVoxFormer.

\section{Related Work}\label{sec:rw}
We relate our work to image-based reconstruction, transformers, and image captioning.

\noindent\textbf{Image-based reconstruction:}
Reconstructing 3D objects using one or multiple image views to a 3D model has been an open research problem for decades. Some approaches use Long Short Term Memory~\cite{lstm} to generate 3D occupancy grids~\cite{3dr2n2}, and Octree~\cite{octree} based deep learning model~\cite{OGN}.
Shen et al.~\cite{frequencyDomain} used images directly by exploiting the Fourier slice theorem~\cite{fourierSlice} as a feature to reconstruct a 3D model. 
Wang et al.~\cite{pix2mesh} used a single image to reconstruct a triangular mesh.

A boundary of a smooth 3D model can be reconstructed by the approach~\cite{occupancy}. The main advantage of this method is a lower memory requirement compared to meshes and voxels. Moreover, Chen et al.~\cite{Chen_2019_CVPR} trained a Generative Adversarial Network (GAN)~\cite{wgan} to learn implicit fields to reconstruct 3D models using images.
Yang et al.~\cite{yang2020robust} used the attention mechanism~\cite{attention} in a feed-forward neural network to reconstruct 3D in voxel from multiple images. Instead of using a Recurrent Neural network (RNN) that suffers from long-term memory loss, Xie et al.~\cite{pix2voxpp} introduced a model with the encoder and decoder in multi- and single-view image reconstruction. Image features are extracted using the encoder and then the decoder then generates coarse 3D volumes.

Yang et al.~\cite{Yang_2021_CVPR} introduced a model that is robust against background noise by using a shape prior. Instead of directly generating 3D volumes, Park et al.~\cite{deepsdf} used a signed distance field (SDF) to generate 3D shapes. Using camera positions and viewing directions, Mildenhall et al.~\cite{nerf} introduced NeRFs that generate a volume density and their spatial radiance. Yang et al.~\cite{Yang_2022_CVPR} also introduced 3D reconstruction from images using estimated camera poses.

Shen et al.~\cite{shen20193d} used a frequency-based lossy compression that can reconstruct about 93\% of the input data. Contrary to this related work, our method uses lossless compression and also encodes it into a 1D structure. SnakeVoxFormer reconstructs 3D voxel models in a space efficient way, by using RLE and the codebook.

\noindent\textbf{Transformer} introduced in NLP~\cite{attention} has been applied in computer vision for many tasks. The state-of-the-art performance in image classification~\cite{vit,swin} is achieved by treating images as tokens that are then fed into various transformer models and object detection in images~\cite{DETR}. Zheng et al.~\cite{SETR} used transformers for image segmentation and it has been shown to classify videos effectively~\cite{arnab2021vivit}.
Liu et al.~\cite{Liu_2022_CVPR} introduced a transformer-based model to classify videos using an inductive bias of locality. 
With notable achievements using transformer network, Wang et al.~\cite{wang2021multi} approached image-to-3D model reconstruction as a sequence-to-sequence problem that transformer scores the state of the art performance. We use the transformer to reconstruct 3D models by mapping the 3D voxel space into a compressed 1D domain.

\noindent\textbf{Image Captioning} adds textual information about the context of an image. Vinyals et al.~\cite{showAndTell} showed how a combination of Convolutional Neural Networks (CNN) and Recurrent Neural Networks (RNNs) is capable of generating captions for a given image. In addition to the CNN and RNN, the attention mechanism has been shown to focus important visual features~\cite{showAndAttend}. 
Li et al.~\cite{imageTransformer} used the transformer-based model for image captioning, and Cornia et al.~\cite{Cornia_2020_CVPR} showed how their meshed attention mechanism can apply a transformer-based model for  image translations. Instead of learning from one feature, Luo et al.~\cite{luo2021dual} use grid and region features to achieve a new state of the art. Better contextual information can be extracted by using region and grid features as opposed to using region features. Our model is different in that it encodes the 3D structure into a 1D lossless compressed data that is highly suitable for deep learning by using the transformer.
\section{Overview}
The input to SnakeVoxFormer (see the overview in Fig.~\ref{fig:overview}) is an RGB image of a single object, and the output is its voxel representation. In the first step, we resize the image to $224\times{2}24$ and input it to the encoder, which is a pre-trained Vision Transformer based on ImageNet in order to  extract image features. The image features are then input to the decoder, which is a transformer with two attention heads that generate the RLE tokens. We then use the dictionary-based encoding (the codebook) (see Tab.~\ref{tab:codebook}) that contains indices to the RLE codes that are then used as reference values. The output of SnakeVoxFormer is a set of tokens that are then used to reconstruct the 3D voxel fully.

\begin{figure*}[hbt]
\centering
\includegraphics[width=0.95\linewidth]{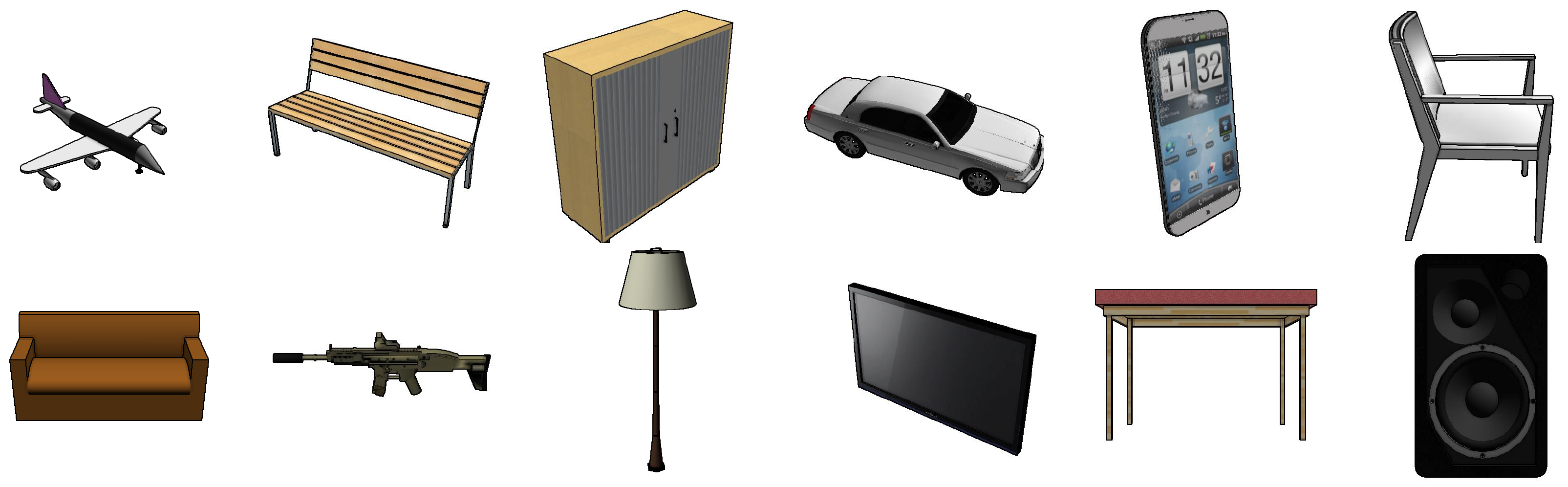}
  \caption{Images from ShapeNet. Top row: airplane, bench, cabinet, car, cellphone, chair.
  Bottom row: couch, firearm, lamp, monitor, table, speaker}\vspace{-3mm}
  \label{fig:inputImage}
\end{figure*}

\section{Data Preprocessing}\label{section:preproc}
We use all 13 categories of objects (airplane, bench, cabinet, car, chair, monitor, lamp, speaker, firearm, couch, table, cellphone, and watercraft)  from the ShapeNet dataset~\cite{shapenet} (see Fig.~\ref{fig:inputImage}). Each object is viewed from 24 different angles. We store the pairs i.e., the image and the voxel representation, for training.

The voxel data is stored in the Binvox file format  $32^3$ (width $\times$ depth $\times$ height). The voxels are sparse, so we compress them in a linear array as RLE (Sec.~\ref{section:RLE}) and we report the compression factor in Tab.~\ref{tab:compressionFactor}.

The RLE-encoded sequences (see Sec.~\ref{section:token}) are analyzed, and repeating sequences are substituted by tokens that are used for the transformer training. Removing the duplicities further reduces the  size of the dataset, which improves learning as compared to learning the raw RLE string with numerous duplicates.

\subsection{Run Length Encoding of Voxels}\label{section:RLE}
RLE~\cite{RLE} stores the data as a sequence of couples of numbers $[rep,val]$, where~$rep$ indicates the number of repetitions and~$val$ the voxel value, so e.g., the couple $[7,0]$ indicates seven consecutive empty voxels. Note that this compression is lossless and symmetric i.e., compression and decompression have the same complexity $\mathcal{O}(n)$, where $n$ is the length of the encoded sequence.
\begin{figure}[hbt]
\centering
\includegraphics[width=0.99\linewidth]{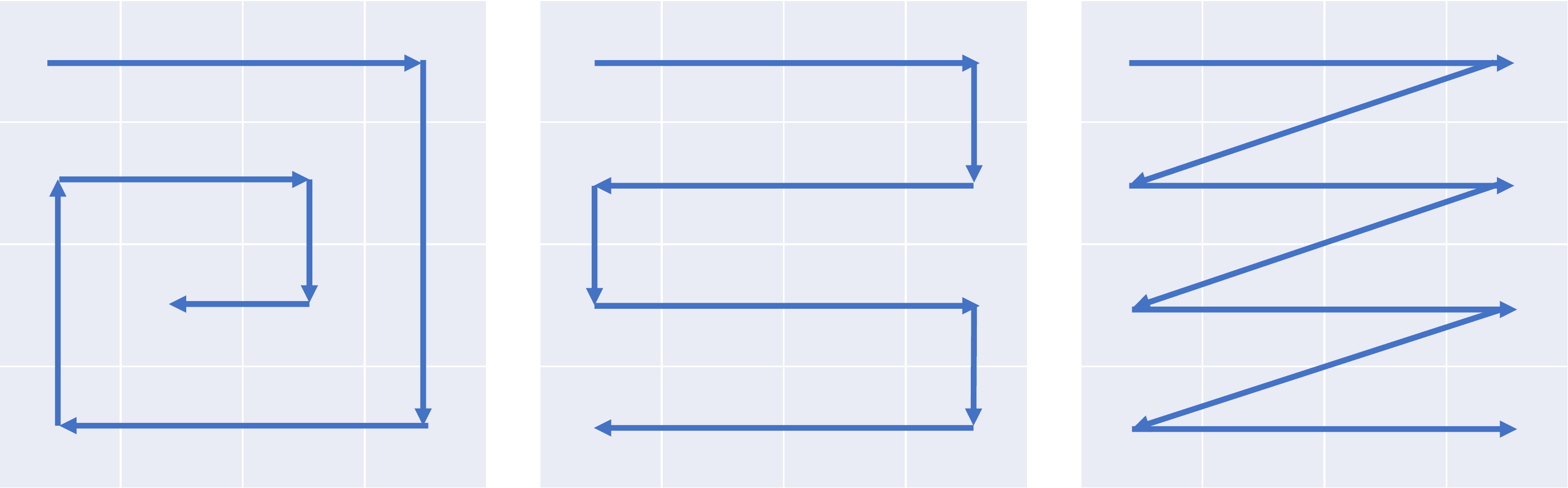}
\caption{Different traversing strategies used for the RLE: a spiral, snake, and raster style scanning.}\vspace{-3mm}  \label{fig:traversal}
\end{figure}

\textbf{Voxel space traversing:} The key to the linearization of the 3D voxel space is its traversing and we have experimented with three different strategies in our approach (Fig.~\ref{fig:traversal}). At each location, the entire array is parsed in the $z$-axis and the voxels are RL encoded. The arrow indicates what is the next column to be encoded.

The first strategy starts on the outer border and spirals down to the center of the voxel space. The second strategy traverses the voxel space like a snake, and the third strategy mimics the scan algorithm of CRT monitors by starting each new line at the beginning. We compare these strategies in Tab.~\ref{tab:RLEcomparison}.

The different traversing strategies depend on the data that they encode. If the scene is almost empty and most of the voxels are centered, the spiral will provide better coherence. 

\textbf{RLE:} We start a string by concatenating a value \textit{E} (empty) and \textit{F} (full) cell. We run RLE to compress the concatenated string so that we actually compress 3D data into 1D array. For example, if an array has values \textit{EEFFF} its RLE will become  \textit{2E3F}. Note that a sequence of \textit{EFEF} will be encoded to \textit{1E1F1E1F} that provides negative compression (inflation). However, these are rare cases and the RLE provides very good compression in most real-world scenarios.

We compute the space efficiency of the RLE by calculating the compression factor $cf$ as 
\begin{equation}
cf=\frac{RLE}{VOX},
\end{equation}
where $RLE$ is the size of the model in RLE representation and $VOX$ is its size in voxels. We report $cf$ of all ShapeNet categories in Tab.~\ref{tab:compressionFactor}. Please note that the $VOX=32^3$ across all objects.
\begin{table}[hbt]
    \centering
    \small
    \begin{tabular}{cccc}
    \toprule
        Category & Snake[B] & Spiral[B] & Raster Scanning[B]\\
    \midrule
        airplane  & 329.91 & 169.89 & 329.91\\
        bench  &  217.63 & 615.06 & 221.37\\ 
        cabinet  &  656.04 & 1428.76 & 660.23\\
        car  &  602.57 & 308.33 & 602.58\\
        chair  &  772.571 & 1063.95 & 774.94\\
        display  &  224.73 & 1446.17 & 229.40\\
        lamp & 600.54 & 589.01 & 600.68\\
        speaker & 820.61 & 1302.15 & 840.17\\
        rifle & 297.88 & 77.24 & 297.88\\
        sofa &  282.63 & 822.60 & 306.49\\
        table &  466.80 & 598.81 & 475.83\\
        telephone & 255.67 & 1114.53 & 256.50\\
        watercraft & 435.09 & 189.73 & 435.10\\
    \bottomrule
        Average &  458.97 & 748.17 & 463.93\\
    \end{tabular}
    \caption{A size of Run Length Encoding per category from Snake, Spiral, Raster scanning traversal.}\label{tab:RLEcomparison}\vspace{-3mm}
\end{table}

\begin{table}[hbt]
    \centering
    \small
    \begin{tabular}{ccc}
    \toprule
        Category & Compression Factor  & RLE Length [B]\\
    \midrule
        airplane  & 0.01007 & 329.91\\
        bench  &  0.00664 & 217.63\\ 
        cabinet  &  0.02002 & 656.04\\
        car  &  0.01839 & 602.57\\
        chair  &  0.02358 & 772.57\\
        display  &  0.00686 & 224.73\\
        lamp & 0.01833 & 600.54\\
        speaker &  0.02504 & 820.61\\
        rifle &  0.00909 & 297.88\\
        sofa &  0.00875 & 286.63\\
        table &  0.01425 & 466.80\\
        telephone &  0.00780 & 255.67\\
        watercraft &  0.01328 & 435.09\\
    \bottomrule
        Overall &  0.01401 & 458.97\\
    \end{tabular}
    \caption{Average of snake traversal compressed factor using RLE and its own length using $32^3$ voxels from ShapeNet.}\label{tab:compressionFactor}\vspace{-3mm}
\end{table}

\subsection{Tokenization}\label{section:token}
Compressing all 3D data into the 1D format improves the space complexity. We compress this further by creating a codebook that stores the RLE sequences as short tokens.

Inspired by Huffman encoding~\cite{huffman1952method}, we find the longest most repeating RLE sequence and substitute it with a single token such as $T_i$, where $i$ is an index of the codebook (see Tab.~\ref{tab:codebook}).
Note, that the transformer's generated tokens are later used to decode their values back to voxels. We use $T_i$ as a key to query the codebook to get $V_i$, a value that corresponds to its original value of RLE compressed data.

The tokens $T_i$ are then used in the decoder (Sec.~\ref{section:decoder}), so it references the codebook to convert it to compressed RLE strings, \textit{$V_i$}, that leads to an occupancy voxel, which corresponds to 3D object to reconstruct. 
\begin{table}[hbt]
    \centering
    \small
    \begin{tabular}{|c|c|}
    \hline
        Token & Value\\
    \hline
        T0  & 2E 3F 1E\\
    \hline
        T1  &  1022E 10F 5E\\ 
    \hline
        T2  &  1E 2F 2E 3F\\
    \hline
        $\dots$ & $\dots$\\
    \hline
    \end{tabular}
    \caption{An example of how the codebook assigns RLE blocks assigned to tokens.}\label{tab:codebook}\vspace{-3mm}
\end{table}

\section{SnakeVoxFormer Architecture and Training}\label{section:architecture}
\subsection{Architecture}
SnakeVoxFormer consists of the encoder  and decoder. The encoder is based on the Vision Transformer~\cite{vit}, which extracts image features inspired by the image captioning style that are then used as input to the decoder to generate its corresponding tokens (Sec.~\ref{section:token}). The tokens are then used to reconstruct the 3D voxels objects using RLE (Sec.~\ref{section:RLE}).

\noindent\textbf{Encoder}\label{section:encoder}
We extract image features using Vision Transformer~\cite{vit} (ViT B/32) pre-trained on ImageNet~\cite{imagenet}. We freeze all the trainable parameters from the ViT to take image features as the encoder of SnakeVoxFormer then we resize images (Fig.~\ref{fig:inputImage}) to $224^2$ in RGB as the input of the encoder.

\noindent\textbf{Decoder}\label{section:decoder}
The decoder takes image features from the encoder and generates tokens $T_i$ (Sec.~\ref{section:token}). We use the original architecture of the transformer~\cite{attention}, but we modified its hyper-parameters, such as the number of heads and the number of hidden units in a feed-forward layer.

\subsection{Training}
We use ShapeNet~\cite{shapenet} dataset and objects categories airplane, bench, cabinet, car, cellphone, chair, couch, firearm, lamp, monitor, table, and speaker which are widely used in a reconstruction problem. The dataset contains 24 different views of each object  and 3D objects are in $32^3$ voxels.

We trained SnakeVoxFormer using CrossEntropy~\cite{crossentropy} as a loss function using Adam~\cite{adam} with default parameters as an optimizer with setting 1e-4 as a learning rate. We set 32, 24, and 96 as the batch size of Snake, Raster Scanning, and Spiral traversal model training.
We validate the model by training four separate models that used 4, 8, 16, and 20 different image views from different angles per object. 

We also trained the three models on different traversing strategies using snake, spiral, and raster scanning.

The model was implemented in Tensorflow and it ran on an Intel i9-12900K clocked at 4.8GHz with Nvidia GeForce RTX 3090. The models were trained for 17 hours (snake traversal), 24 hours (spiral), and 143.5 hours (raster scanning). The different training time comes from different batch size and the size of the used dataset. We did not observe a notable performance difference after 75 hours of training on the raster scanning model.

We will provide the source code of SnakeVoxFormer.

\section{Experiments and Results}
\begin{table}[hbt]
    \centering
    \small
    \begin{tabular}{cccc}
    \toprule
        Input Feature & Feed Forward & Heads & Layers\\
    \midrule
        32 & 64 & 8 & 6\\
        64 & 128 & 8 & 8\\ 
        128 & 256 & 8 & 8\\
        128 & 256 & 8 & 6\\
        128 & 256 & 2 & 8\\
        256 & 512 & 2 & 6\\
        256 & 256 & 4 & 8\\
        512 & 1024 & 2 & 4\\
        512 & 256 & 2 & 12\\
    \bottomrule
        \underline{768} & \underline{256} & \underline{2} & \underline{3}
    \end{tabular}
    \caption{A set of parameters for SnakeVoxFormer}\label{tab:ablation}\vspace{-3mm}
\end{table}

\subsection{Ablation}
We performed several ablation experiments and  decided to use three layers, 768 as the size of input features and 256 as the dimension of the feed-forward network. We use two heads in the multi-head-attention layers. We set the dropout rate to 0.2 to avoid over-fitting (see Tab.~\ref{tab:ablation}).

We perform an extensive parameter search to provide the best performance. We found that  if the number of heads is greater than four, models start to generate incorrect results and it does not even perform well on the training data. We also test it by increasing the number of layers, but the model started to overfit the training data quickly.

\subsection{Metric}
We tested the four trained models for 4, 8, 16, and 20 different view images of a single object. We computed the performance based on Intersection over Union (IoU) between the ground truth voxels and reconstructed voxels using the trained models. A higher IoU corresponds to a better reconstruction performance. We define IoU as:
\begin{equation}
    IoU = 
    \frac{\Sigma_{(i,j,k)} I(y(i,j,k))I(\hat{y}(i,j,k))}{\Sigma_{(i,j,k)}I(I(y(i,j,k))+I(\hat{y}(i,j,k))  )}
\end{equation}
where, $y(i,j,k)$ is a generated occupancy voxel and $\hat{y}(i,j,k)$ is a ground truth voxel and $I(\cdot)$ represent an indicator function.

We used 130,060, 399,380, and 78,000 images to train models using Snake, Raster Scanning, and Spiral indexing traversal for 13 categories.

We tested our models from 1,300 images from unseen data during the training process and report average IoU values per different views in Tab.~\ref{tab:comparison}. We also provide detailed IoU values for each category from models that are trained on different views in Tab.~\ref{tab:detailedMetricSnake}, Tab.~\ref{tab:detailedMetricRaster} and Tab.~\ref{tab:detailedMetricSpiral}.

\begin{table}[hbt]
    \centering
    \small
    \begin{tabular}{lcccc}
    \toprule
        IoU/Views & 4 & 8 & 16 & 20 \\
    \midrule
        3D-R2N2~\cite{3dr2n2}  & 0.625 & 0.635 & 0.636 & 0.636\\
        AttSets~\cite{yang2020robust} &  0.675 & 0.685 & 0.688 & 0.693 \\ 
        Pix2Vox-A~\cite{pix2vox}  &  0.697 & 0.702 & 0.705 & 0.706\\
        Pix2Vox++/A~\cite{pix2voxpp}  &  0.708 & 0.715 & 0.718 & 0.719\\
        VoIT~\cite{wang2021multi}  &  0.605 & 0.681 & 0.706 & 0.711\\
        VoIT+~\cite{wang2021multi}  &  0.695 & 0.707 & 0.714 & 0.715\\
        EVoIT~\cite{wang2021multi} &  0.609 & 0.698 & \underline{0.729} & \underline{0.735}\\
        TMVNet~\cite{TMVNet} & \underline{0.718} & \underline{0.719} & 0.721 & - \\
    \bottomrule
        Ours (Snake) &  \textbf{0.843} & \textbf{0.874} & \textbf{0.904} & 0.912\\
        Ours (Raster) &  0.746 & 0.834 & 0.865 & 0.861\\
        Ours (Spiral) &  0.829 & 0.864 & 0.889 & \textbf{0.933}\\
    \end{tabular}
    \caption{Mean IoU metrics for 13 categories.}\label{tab:comparison}\vspace{-3mm}
\end{table}
\subsection{Result}
\begin{figure*}[hbt]
\centering
\includegraphics[width=0.95\linewidth,height=\textheight,keepaspectratio]{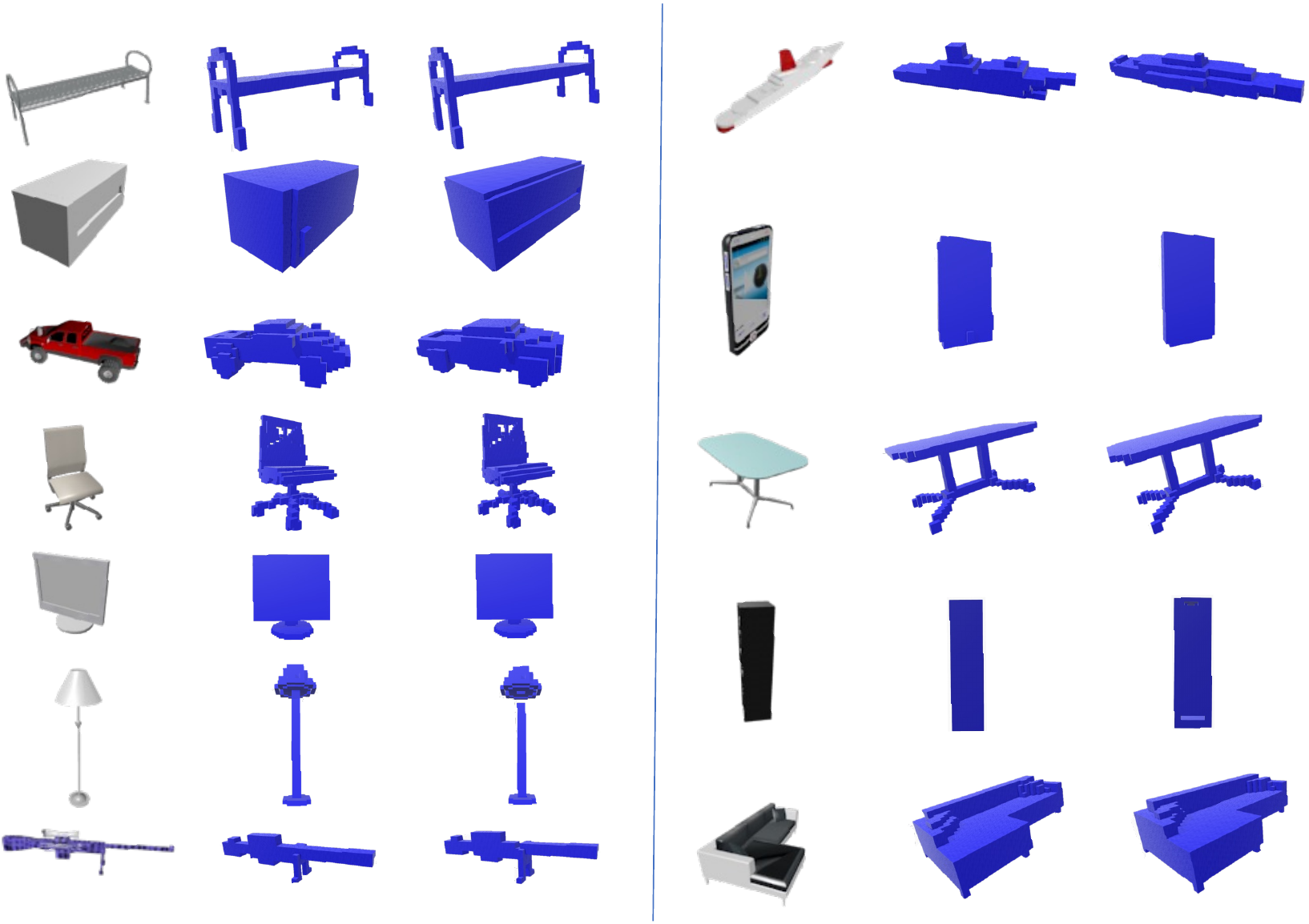}
  \caption{3D reconstruction results on ShapeNet from our trained model. The left column is the input image, the middle column is the ground truth voxel and the right column shows our reconstructed  voxel.}\vspace{-3mm}
  \label{fig:generation_result}
\end{figure*}

Figure.~\ref{fig:generation_result} shows examples of 3D voxels generated by SnakeVoxFormer and Table~\ref{tab:comparison} shows the comparison to eight models include 3D-R2N2~\cite{3dr2n2}, AttSets~\cite{yang2020robust}, Pix2Vox-A~\cite{pix2vox}, Pix2Vox++/A~\cite{pix2voxpp}, VoIT~\cite{wang2021multi}, VoIT+~\cite{wang2021multi}, EVoIT~\cite{wang2021multi}, and TMVNet~\cite{TMVNet}.
We underline the state-of-the-art performance of the previous works and we put our performance in bold text at the bottom line. 

Table.~\ref{tab:comparison} shows that our models have a higher mean IoU value compared to previous works.

We compare the snake and spiral traversal strategies in Tab.~\ref{tab:comparison}. We show the percentage differences between our metrics and the best values from the previous works. Our snake model scores  12.5\%, 15.5\%, 17.5\%, and 17.7\% better than TMVNet~\cite{TMVNet}, TMVNet~\cite{TMVNet}, EVoIT~\cite{wang2021multi}, EVoIT~\cite{wang2021multi} in 3D reconstruction using 4, 8, 16, 20 views reconstruction respectively.
\begin{table}[hbt]
    \centering
    \small
    \begin{tabular}{cllll}
    \toprule
        Category(IoU) & 4 Views& 8 Views& 16 Views& 20 Views\\
    \midrule
        airplane  & 0.7506 & 0.7882 & 0.8805 & 0.8786\\
        bench  &  0.8694 & 0.8960 & 0.9311 & 0.9349\\ 
        cabinet  &  0.9095 & 0.9511 & 0.9621 & 0.9531\\
        car  &  0.9074 & 0.9411 & 0.9582 & 0.9671\\
        chair  &  0.8521 & 0.8860 & 0.9348 & 0.9327\\
        display  &  0.7551 & 0.8210 & 0.8576 & 0.9092\\
        lamp &  0.8658 & 0.8756 & 0.8718 & 0.8746\\
        speaker &  0.8240 & 0.8588 & 0.9010 & 0.9321\\
        rifle &  0.7276 & 0.7538 & 0.7838 & 0.7875\\
        sofa &  0.8930 & 0.9067 & 0.9025 & 0.9444\\
        table &  0.8844 & 0.9221 & 0.9249 & 0.9607\\
        telephone &  0.9070 & 0.9296 & 0.9480 & 0.9044\\
        watercraft &  0.8173 & 0.8255 & 0.8896 & 0.8742\\
    \bottomrule
        Overall &  0.8433 & 0.8735 & 0.9035 & 0.9118\\
    \end{tabular}
    \caption{IoU metrics per category using Snake Traversal method.}\label{tab:detailedMetricSnake}\vspace{-3mm}
\end{table}

\begin{table}[hbt]
    \centering
    \small
    \begin{tabular}{cllll}
    \toprule
        Category(IoU) & 4 Views& 8 Views& 16 Views& 20 Views\\
    \midrule
        airplane  & 0.660 & 0.793 & 0.822 & 0.798\\
        bench  &  0.776 & 0.840 & 0.875 & 0.911 \\ 
        cabinet  &  0.816 & 0.875 & 0.910 & 0.896\\
        car  &  0.757 & 0.855 & 0.883 & 0.871\\
        chair  &  0.724 & 0.833 & 0.868 & 0.858\\
        display  &  0.646 & 0.790 & 0.816 & 0.805\\
        lamp &  0.823 & 0.897 & 0.934 & 0.940\\
        speaker &  0.771 & 0.867 & 0.887 & 0.893\\
        rifle &  0.630 & 0.731 & 0.743 & 0.706\\
        sofa &  0.801 & 0.849 & 0.897 & 0.879\\
        table &  0.783 & 0.864 & 0.882 & 0.924\\
        telephone &  0.811 & 0.863 & 0.886 & 0.888\\
        watercraft &  0.695 & 0.788 & 0.845 & 0.826\\
    \bottomrule
        Overall &  0.746 & 0.834 & 0.865 & 0.861\\
    \end{tabular}
    \caption{IoU metrics per category using Raster Scanning Traversal method.}\label{tab:detailedMetricRaster}\vspace{-3mm}
\end{table}

\begin{table}[hbt]
    \centering
    \small
    \begin{tabular}{cllll}
    \toprule
        Category(IoU) & 4 Views& 8 Views& 16 Views& 20 Views\\
    \midrule
        airplane  & 0.737 & 0.796 & 0.829 & 0.907\\
        bench  &  0.727 & 0.844 & 0.877 & 0.933 \\ 
        cabinet  &  0.893 & 0.886 & 0.915 & 0.957\\
        car  &  0.917 & 0.936 & 0.923 & 0.970\\
        chair  &  0.816 & 0.851 & 0.860 & 0.931\\
        display  &  0.842 & 0.818 & 0.869 & 0.899\\
        lamp &  0.896 & 0.914 & 0.939 & 0.957\\
        speaker &  0.891 & 0.895 & 0.931 & 0.927\\
        rifle &  0.790 & 0.793 & 0.837 & 0.887\\
        sofa &  0.778 & 0.869 & 0.897 & 0.983\\
        table &  0.807 & 0.890 & 0.915 & 0.920\\
        telephone &  0.855 & 0.895 & 0.894 & 0.917\\
        watercraft &  0.883 & 0.844 & 0.871 & 0.947\\
    \bottomrule
        Overall &  0.829 & 0.864 & 0.889 & 0.933\\
    \end{tabular}
    \caption{IoU metrics per category using Spiral Traversal method.}\label{tab:detailedMetricSpiral}\vspace{-3mm}
\end{table}

\section{Conclusion}
\begin{figure}[hbt]
\centering
\includegraphics[width=0.99\linewidth,height=0.5\textheight,keepaspectratio]{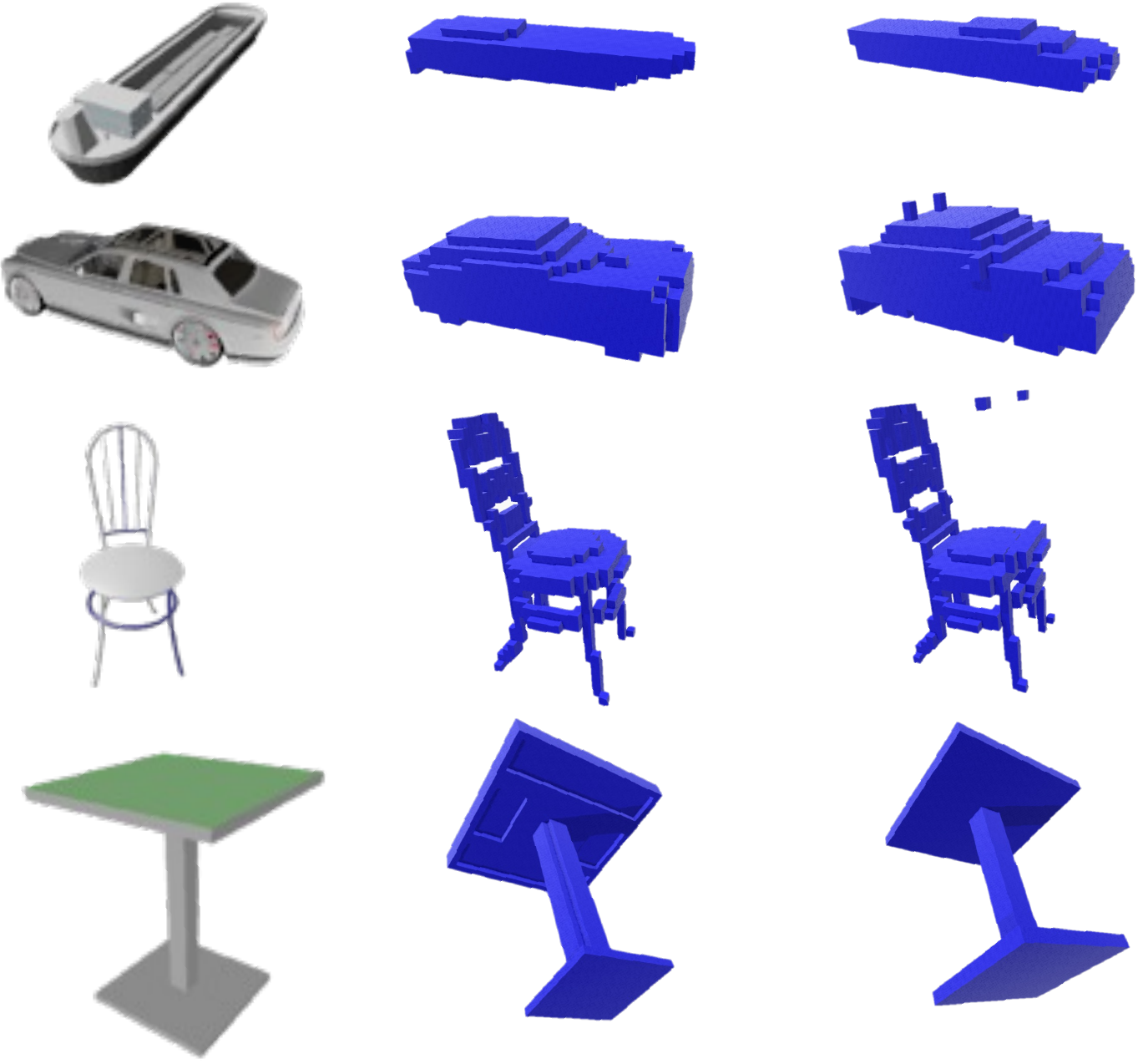}
  \caption{Our model fails on small parts of objects. Input image (left), ground truth (middle), SnakeVoxFormer (right).}\vspace{-3mm}
  \label{fig:failure}
\end{figure}
We introduced SnakeVoxFormer, a transformer-based 3D reconstruction that traverses 3D space and Run Length Encodes the voxel. Our algorithm uses lossless compression of the ShapeNet dataset to only around 1\% of its original size while encoding the RLE chunks into tokens suitable for NLP-based training. Moreover, we show that SnakeVoxFormer outperforms the new state-of-the-art algorithms for multi-view reconstruction from 2D images to voxels.

One of the limitations of our model is that RLE assumes non-noisy structures that make SnakeVoxFormer suitable for man-made objects. Also, our method fails on small structures such as sharp edges and small object parts. Another limitation is that it is suitable only for individual objects (see Fig.~\ref{fig:failure}).

Possible future work includes exploring different traversing strategies and their effect on the overall model performance. It would also be interesting to see how our method would generalize for the entire scene.

We will provide the source code of SnakeVoxFormer.
% %------------------------------------------------------------------------

{\small
\bibliographystyle{ieee_fullname}
\bibliography{main}
}

\end{document}